\documentclass[journal]{IEEEtran}

\ifCLASSINFOpdf
\else
   \usepackage[dvips]{graphicx}
\fi
\usepackage{url}

\usepackage{graphicx}
\usepackage{cite}
\usepackage{dsfont}
\usepackage{bbm}
\usepackage{float}
\usepackage{booktabs}
\usepackage{graphicx}
\usepackage{multirow}
\usepackage{tabularx}
\usepackage{hyperref}
\usepackage{amsmath}
\usepackage{balance}
\usepackage{orcidlink}
\usepackage{pifont} 
\hypersetup{colorlinks=true, linkcolor=blue, anchorcolor=blue, citecolor=blue}

\begin{document}
\title{Exploring Token-Level Augmentation in Vision Transformer for Semi-Supervised Semantic Segmentation}

\author{Dengke Zhang~\orcidlink{0009-0001-2941-0084}, Quan Tang~\orcidlink{0000-0003-4011-6166}, Fagui Liu~\orcidlink{0000-0003-1135-4982}, \IEEEmembership{Member, IEEE}, Haiqing Mei, C. L. Philip Chen~\orcidlink{0000-0001-5451-7230}, \IEEEmembership{Fellow, IEEE}
\thanks{This work was partially supported by the Guangdong Major Project of Basic and Applied Basic Research (2019B030302002), the Major Key Project of PCL (PCL2023A09), the Science and Technology Major Project of Guangzhou (202007030006), and the Science and Technology Project of Guangdong Province (2021B1111600001). Codes will be available. \textit{(Corresponding authors: Quan Tang and Fagui Liu.)}}
\thanks{D. Zhang and F. Liu are with the School of Computer Science and Engineering, South China University of Technology, Guangzhou 510006, China. F. Liu is also with the Department of New Network, Pengcheng Laboratory, Shenzhen 518000, China (e-mail: fgliu@scut.edu.cn).

Q. Tang is with the Department of New Network, Pengcheng Laboratory, Shenzhen 518000, China (e-mail: tangq@pcl.ac.cn).

% F. Liu is with the School of Computer Science and Engineering, South China University of Technology, Guangzhou 510006, China, and also with the Department of New Network, Pengcheng Laboratory, Shenzhen 518000, China (e-mail: fgliu@scut.edu.cn).

H. Mei is with the China Tobacco Guangdong Industrial Corporation, Guangzhou 510385, China

C. L. P. Chen is with the School of Computer Science and Engineering, South China University of Technology, Guangzhou 510006, China, and State Key Laboratory of Management and Control for Complex Systems, Institute of Automation, Chinese Academy of Sciences, Beijing 100080, China.
}}

\markboth{Journal of \LaTeX\ Class Files}
{Shell \MakeLowercase{\textit{et al.}}: Bare Demo of IEEEtran.cls for IEEE Journals}
\maketitle

\begin{abstract}
Semi-supervised semantic segmentation has witnessed remarkable advancements in recent years. However, existing algorithms are based on convolutional neural networks and directly applying them to Vision Transformers poses certain limitations due to conceptual disparities. To this end, we propose TokenMix, a data augmentation technique specifically designed for semi-supervised semantic segmentation with Vision Transformers. TokenMix aligns well with the global attention mechanism by mixing images at the token level, enhancing learning capability for contextual information among image patches. We further incorporate image augmentation and feature augmentation to promote the diversity of augmentation. Moreover, to enhance consistency regularization, we propose a dual-branch framework where each branch applies image and feature augmentation to the input image. We conduct extensive experiments across multiple benchmark datasets, including Pascal VOC 2012, Cityscapes, and COCO. Results suggest that the proposed method outperforms state-of-the-art algorithms with notably observed accuracy improvement, especially under limited fine annotations.
\end{abstract}


\IEEEpeerreviewmaketitle

\section{Introduction}
\IEEEPARstart{D}{eep} network advances have significantly improved fully-supervised semantic segmentation\cite{1},\cite{2},\cite{3},\cite{bsiou},\cite{4}, but these methods heavily rely on large-scale finely annotated datasets, demanding considerable human resources. Semi-supervised semantic segmentation (S4) mitigates this by reducing the need for annotated data, allowing effective learning from limited labels. S4 techniques can largely be attributed to two major paradigms: self-training\cite{5},\cite{6},\cite{7}.\cite{8} and consistency regularization\cite{9},\cite{10},\cite{11},\cite{12},\cite{13}. Self-training achieves knowledge transfer from labeled data to unlabeled data through iterative cycles. Consistency regularization promotes model learning through consistency constraints of the same image under different augmentation views.

Recent research in S4 is mostly constructed upon Convolutional Neural Network (CNN)\cite{15}. Although CNN has shown excellent performance in visual tasks, Vision Transformer (ViT)\cite{16}, as an emerging vision representation learning model, has demonstrated its superior performance and unique advantages in multiple computer vision tasks\cite{4},\cite{laformer},\cite{cross-vit},\cite{detr}. However, algorithms developed for CNN work less effectively on ViT \cite{14} due to their inherent architectural and conceptual disparities. Therefore, exploring how to effectively integrate ViT into S4 tasks to leverage its powerful global information processing capabilities requires in-depth research. 
% To this end, we propose a new data augmentation technique called TokenMix, and further expand augmentation space by introducing randomness in the high-level feature space, i.e., feature augmentation. Furthermore, a dual-branch framework is used to complement each other and enhance consistency regularization, achieving higher segmentation accuracy even with limited supervision.

% CutMix\cite{17}, by cutting and pasting parts of other images onto the original image, increases data diversity and forces the model to learn global information. However, its advantages may be diminished in ViT, as ViT inherently excel at capturing long-range dependencies and global context. In contrast to this, TokenMix facilitates a better understanding of global information through token-level exchanges. Moreover, in semantic segmentation task, the information at the edges cut by CutMix may be discontinuous, leading to the blurring of important detail features, which can interfere with the model's learning. TokenMix, on the other hand, operates on tokens at a higher level of abstraction. It helps to maintain the coherence and naturalness of the image content when mixing information from different images, reducing the interference with the model learning. 

CutMix\cite{17} swaps regions from different images at the pixel level, serving as a common augmentation method in CNN-based S4 to introduce more diverse views of images. However, ViT divides an image into patches, treating each as an independent token for modeling. CutMix's design isn't fully compatible with ViT's patch-based approach, which may cause information loss or inconsistencies. For example, CutMix creates boundary discontinuities that distort critical details. This prevents ViT from accurately capturing original semantic features in image patches. To prevent the above issue, we propose TokenMix, which operates on tokens and aligns with ViT's patch-based modeling. TokenMix helps to maintain the coherence and naturalness of the image content while introducing new contextual information by mixing tokens from different images. This helps ViT better understand contextual information among patches and improves the utilization efficiency of unlabeled data.

\begin{figure*}[htbp]
\centerline{\includegraphics[height=2in]{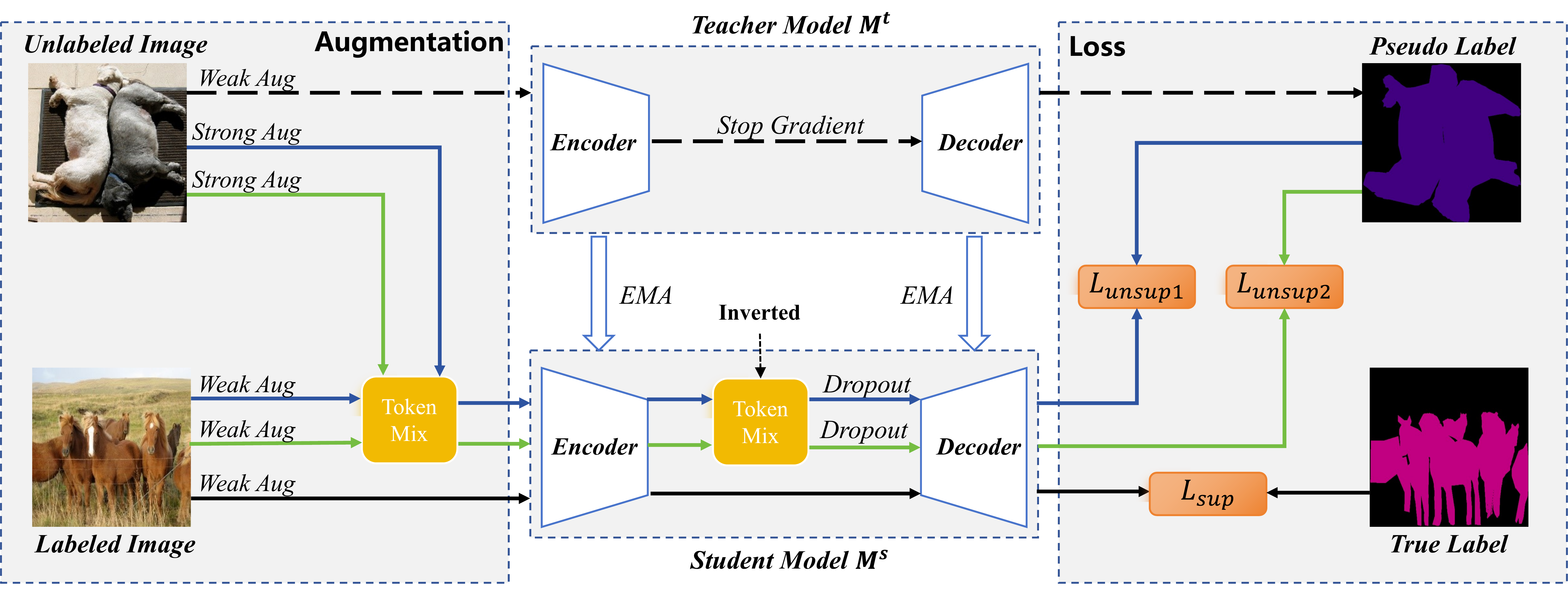}}
\caption{The proposed framework. Labeled images undergo weak augmentation and are fed into the student model to generate predictions, which are then used to compute the supervised loss against finely annotated labels. Blue and green lines denote two independent branches for unlabeled images. Unlabeled images undergo weak augmentation and are input into the teacher model to generate pseudo-labels. Subsequently, the same unlabeled images undergo strong augmentation, TokenMix, and Dropout within the student model to calculate the unsupervised loss using the pseudo-labels.}
\label{fig1}
\end{figure*}

Weak image augmentations (e.g., flipping) and strong image augmentations (e.g., image blur) capitalize on the advantages of consistency regularization in S4, yet their application is limited to the image level. To expand the augmentation space, Unimatch\cite{9} uses a separate Dropout branch as a supplement, which is very effective on CNN-based models but performs poorly on  ViT-based models\cite{13},\cite{14}. The reason lies in ViT's inherently greater optimization difficulty. When using various augmentation techniques in a multi-branch framework, the inconsistencies caused by different branches exacerbate the optimization challenge, especially when few labels provide strong guidance. In light of this, we use two consistent branches that simultaneously employ image and feature augmentation to expand the augmentation space and enhance consistency regularization while mitigating the performance degradation caused by optimization difficulty. 

Armed with the proposed mechanisms, our method effectively harnesses ViT's global modeling capacity and fully taps into its potential within the domain of S4. The main contributions are summarized as follows:

\begin{itemize}
\item[1)] We propose TokenMix that mixes images at the token level, which aligns well with ViT and introduces new contextual information to improve understanding of patch relationships and enhance utilization of unlabeled data.

\item[2)] We present a dual-branch framework where both branches use image and feature augmentation to expand augmentation space and promote consistency regularization.

\item[3)] Extensive experiments on PASCAL VOC 2012, Cityscapes, and COCO datasets demonstrate competitive performance of the proposed method, especially when finely annotations are rare.
\end{itemize}

\section{PROPOSED METHOD}
\subsection{Overall Framework}
In S4 task, the dataset is divided into labeled and unlabeled datasets. The labeled dataset $D_l=\{(x_i^l,y_i^l)\}_{i=1}^{N_l}$ consists of $N_l$ images $x^l\in{R^{H\times W\times 3}}$ and corresponding labels $y^l\in{R^{H\times W}}$, where $H$ and $W$ represent the height and width, respectively, and $y^l$ are integers within the range $[0, C)$, with $C$ being the number of classes. The unlabeled dataset $D_u=\{x_i^u\}_{i=1}^{N_u}$ consists of $N_u$ images $x^u\in{R^{H\times W\times 3}}$ and $N_u \gg N_l$.

Fig. \ref{fig1} presents the proposed dual-branch S4 framework, which is developed based on the Teacher-Student\cite{26} paradigm. It comprises two networks with identical architectures: the teacher $M^t$ and the student $M^s$. The student model is updated through gradient descent, while the teacher model is updated using Exponential Moving Average (EMA) to get more stable and reliable pseudo labels:
\begin{equation}
M_i^t=\theta \times M_{i-1}^t + (1 - \theta) \times M_{i-1}^s,
\end{equation}
where $ M_{i}^t$ and $ M_{i}^s$ represent the model parameters at the i-th iteration of the training, and $\theta$ denotes the momentum decay factor that signifies the update rate. The loss comprises supervised loss and the unsupervised loss: $L=L_{sup} + L_{unsup}$. The labeled data $x^l$, after undergoing weak augmentation $\alpha$, is fed into the student model $M^s$ to obtain the segmentation prediction $p^l=M^s(\alpha(x^l))$, which is then used to calculate the supervised loss $L_{sup}$ against the label $y^l$:
\begin{equation}
L_{sup}=l_{CE}(p^l, y^l),
\end{equation}

Unlabeled data first undergoes weak augmentation, and then it is input into the teacher model to obtain the segmentation result $y^u=\mathop{\arg\max}\limits_{c}M^t(\alpha(x^u))$, which is used as pseudo-label. Subsequently, the result is compared with the segmentation outcome $p^u=M^s(A(\alpha(x^u))$ obtained by the student model after the data undergoes weak augmentation and strong augmentation $A$ to calculate the unsupervised loss:
\begin{equation}
L_{unsup} = l_{CE}(p^u, y^u),
\end{equation}

To ensure that the teacher model can provide reliable guidance, only the losses of pixels whose confidences are greater than predefined threshold $\rho$ are calculated:
\begin{equation}
L_{unsup} = L_{unsup} \odot \mathbbm{1}(Conf > \rho)
\end{equation}
where $Conf=\mathop{\max}\limits_{c}(softmax(M^t(\alpha(x^u))))$ is the confidence map and $\odot$ represents element-wise multiplication.

\begin{figure}[htbp]
\centerline{\includegraphics[width=\columnwidth]{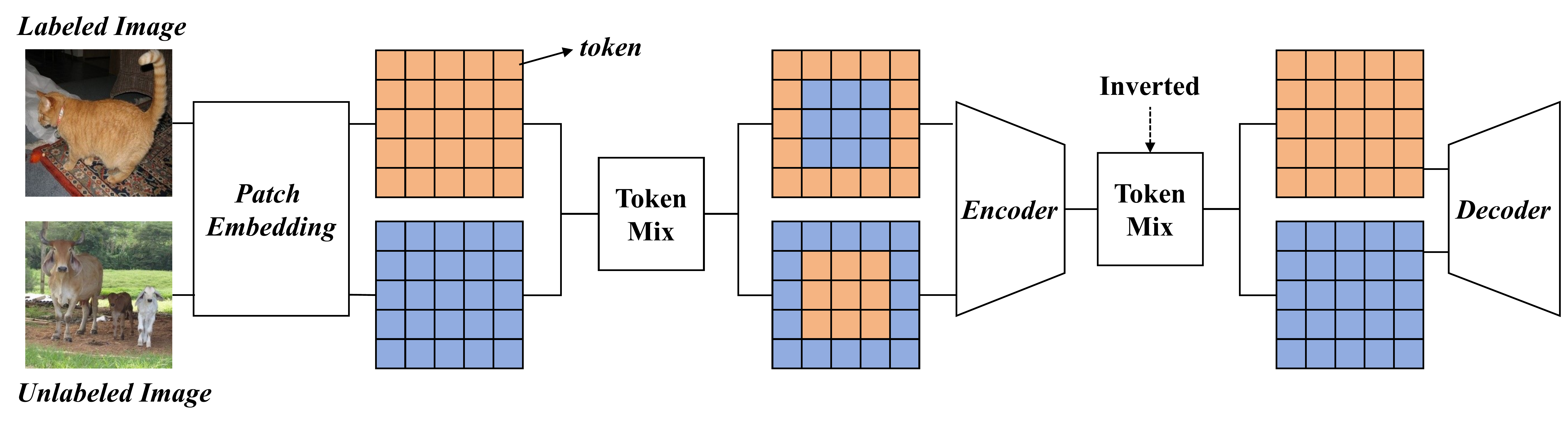}}
\caption{Process of TokenMix. Tokens are mixed subsequent to the patch embedding and are reverted to their initial arrangement before the decoder.}
\label{fig_tokenmix}
\end{figure}

\subsection{TokenMix}

The design of TokenMix is more in line with ViT's processing procedure. Its operation at the token level can better work with the self-attention mechanism, using ViT's advantages to understand the relationship between mixed tokens.

The structure of model $M$ can be divided into three parts: Patch Embedding, Encoder, and Decoder. Patch Embedding divides the image $x$ into patches and transforms them into high-dimensional tokens $g$. Encoder performs positional encoding on the tokens and then extracts image features $f$ through self-attention and feedforward layers. Decoder calculates the final segmentation results $p$ of the image based on the features extracted by the Encoder. The process is shown in Fig. \ref{fig_tokenmix}.

TokenMix first exchanges some tokens between Patch Embedding and Encoder, and then exchanges these tokens back between Encoder and Decoder. TokenMix is carried out between labeled and unlabeled images. This allows the model to indirectly apply the knowledge obtained from supervised learning on unlabeled data, helping the model learn more accurate feature representations. Assuming that labeled and unlabeled images are transformed into tokens $g^l$ and $g^u$ after passing through Patch Embedding, the operation of the first token exchange can be represented as:
\begin{equation}
\left\{  
     \begin{array}{lr}  
     g^u = g^u \odot (1 - m) + g^l \odot m, & \\  
     g^l = g^l \odot (1 - m) + g^u \odot m, & \\   
     \end{array}  
\right. 
\end{equation}
where $m$ is a binary mask where 1 indicates token exchange and 0 indicates no change. After $g^l$ and $g^u$ are processed by Encoder, tokens are then exchanged back:
\begin{equation}
f^u=f^u \odot (1 - m) + f^l \odot m,
\end{equation}

% We do not use TokenMix on labeled images as this may affect the learning of annotation information, especially when the labels are scarce.

\subsection{Dual-Branch}
During training, each iteration with Dropout can be seen as evaluating the same input data under different sub-networks. This mechanism encourages the model to explore broader augmentation space. At the same time, these sub-networks are expected to maintain consistency in predictions for unlabeled data, thereby implicitly enforcing consistency regularization. Dropout is utilized between the Encoder and Decoder:
\begin{equation}
f^u=Dropout(f^u),
\end{equation}

Integrating the aforementioned augmentation methods, we present a dual-branch framework, where both branches apply a series of weak augmentation, strong augmentation, TokenMix, and Dropout to unlabeled data, as shown in Fig. \ref{fig1}. Consequently, the unsupervised loss is composed of two parts:
\begin{equation}
L_{unsup} = (L_{unsup1} + L_{unsup2}) \times 0.5,
\end{equation}

Two branches receive similar but not identical inputs, and encouraging them to generate consistent outputs can serve as a form of consistency regularization. 
% Moreover, the two branches' learning effects can suppress errors' propagation. If the predictions from both branches are consistent, they have jointly identified stable features in the data, and the model's parameter updates normally like a single branch. Conversely, if the predictions are inconsistent, which implies the potential for errors, the model parameter updates will tend to reduce these discrepancies, thereby suppressing the propagation of errors to a certain extent.

\section{EXPERIMENTS}
\subsection{Experiment Setup}
\textit{Datasets:}We evaluate different methods on three datasets: \textbf{Pascal VOC 2012} \cite{36} (1,464 training and 1,449 validation images, 21 categories, augmented with 9,118 coarsely labeled SBD training images\cite{37}); \textbf{Cityscapes} \cite{38} (19 categories, 2,975 training and 500 validation images); and \textbf{COCO} \cite{39} (81 categories, 118,000 training and 5,000 validation images).

\textit{Implementation Details:} For fair comparison, all methods use the Segmenter-S model\cite{32} with SGD optimizer whose momentum is 0.0001. Learning rates are 0.001 (Pascal), 0.005 (Cityscapes), and 0.004 (COCO), with a batch size of 16 (8 labeled, 8 unlabeled). Crop sizes are 320$\times$320 (Pascal, COCO) and 512$\times$512 (Cityscapes). Training durations under a poly learning rate scheduler are 80, 240, and 10 epochs, respectively. Performance is evaluated using mean Intersection over Union (mIoU).

\subsection{Comparisons With State-of-the-Arts}
Following previous methods, we replicate the labeled data to match the quantity of unlabeled data. Sup Only refers to the baseline approach, where the model is trained using only labeled data. The comparison between other methods and the baseline can reflect the efficiency of utilizing unlabeled data. The heading of Table \ref{sota} denotes the number of labeled images.

Our method performs well across multiple datasets (Table \ref{sota}). On Pascal, our approach outperforms existing methods under all partition protocols. On Cityscapes, our method achieves significant gains, particularly in scenarios with limited labeled data. On the challenging COCO dataset, our method shows notable improvements, further demonstrating our method's superiority in situations where the label ratio is low.

\setlength{\floatsep}{5pt} % 设置浮动体之间的间距
\setlength{\textfloatsep}{5pt} % 设置浮动体与文本之间的间距
\setlength{\intextsep}{5pt} % 设置嵌入文本中的浮动体与文本的间距

\begin{table}[htbp]
\renewcommand{\arraystretch}{0.8}
\caption{Comparison on Pascal, Cityscapes and COCO}
\centering
\begin{tabularx}{1\linewidth}{p{2cm} >{\centering\arraybackslash}X >{\centering\arraybackslash}X >{\centering\arraybackslash}X >{\centering\arraybackslash}X}
\toprule
Pascal&732(1/2)&366(1/4)&183(1/8)&92(1/16) \\
\midrule
Sup Only&72.92&69.73&66.44&60.12 \\
% CutMix&76.17&74.72&72.57&70.84&67.24 \\
AugSeg\cite{10}&76.41&74.94&74.43&68.84 \\
UniMatch\cite{9}&76.17&74.90&71.53&63.61 \\
AllSpark\cite{14}&74.85&72.20&71.89&68.02 \\
% TokenMix&77.31&76.83&75.85&75.24&71.00 \\
Ours&\textbf{77.35}&\textbf{76.12}&\textbf{75.40}&\textbf{72.90} \\

\midrule
Cityscapes&1488(1/2)&744(1/4)&372(1/8)&186(1/16) \\
\midrule
Sup Only&72.56&70.90&66.82&61.61 \\
AugSeg\cite{10}&74.54&73.51&72.31&69.01 \\
UniMatch\cite{9}&\textbf{74.60}&73.30&72.22&69.45 \\
AllSpark\cite{14}&72.94&71.87&68.27&66.92\\
Ours&74.23&\textbf{73.62}&\textbf{72.65}&\textbf{70.85} \\

\midrule
COCO&1849(1/64)&925(1/128)&463(1/256)&232(1/512) \\
\midrule
Sup Only&36.57&33.67&27.10&16.56 \\
AugSeg\cite{10}&39.58&37.40&31.33&27.79 \\
UniMatch\cite{9}&39.98&35.97&31.07&24.60 \\
Ours&\textbf{43.91}&\textbf{38.58}&\textbf{34.96}&\textbf{29.84} \\

\bottomrule
\label{sota}
\end{tabularx}
\end{table}

\begin{table}[htbp]
\caption{Comparison of different augmentations on Pascal}
  \renewcommand{\arraystretch}{0.8}
\centering
\begin{tabularx}{1\linewidth}{p{5cm} >{\centering\arraybackslash}X >{\centering\arraybackslash}X >{\centering\arraybackslash}X}
\toprule
Augmentation&366&183&92 \\
\midrule
CutMix&72.57&70.84&67.24 \\
Adaptive CutMix\cite{10}&74.94&74.43&68.84 \\
Adaptive CutMix + TokenMix(Ours)&75.16&74.60&70.09 \\
ClassMix\cite{40}&74.76&74.20&70.57 \\
ClassMix + TokenMix(Ours)&75.51&74.88&70.10 \\
\textit{TokenMix*}\cite{tokenmix}&75.06&74.79&69.18 \\
\textbf{TokenMix(Ours)}&\textbf{75.85}&\textbf{75.24}&\textbf{71.00} \\
\bottomrule
\label{augmentation}
\end{tabularx}
\end{table}

\begin{table}
  \caption{Integrating TokenMix into other methods on Pascal (366 labels).}
  \label{integration}
  \centering
  \renewcommand{\arraystretch}{0.8}
  % 定义新的列类型 C，用于居中对齐
  \newcolumntype{C}{>{\centering\arraybackslash}X}
  \setlength{\tabcolsep}{-4pt} % 默认是 6pt
  \begin{tabularx}{\linewidth}{p{1.45cm}<{\centering}|CCCp{1.5cm}}
    \toprule
    TokenMix & CorrMatch\cite{corrmatch} & RankMatch\cite{rankmatch} & PrevMatch\cite{prevmatch} & UCCL\cite{uccl} \\
    \midrule
    \ding{56} & 74.60 & 74.59 & 75.38 & 74.59 \\
    \ding{52} & 75.60 & 75.24 & 76.02 & 75.59 \\
    \bottomrule
  \end{tabularx}
\end{table}

\begin{table}[htbp]
\caption{Comparison of different branch designs on Pascal}
  \renewcommand{\arraystretch}{0.8}
\centering
\begin{tabularx}{1\linewidth}{p{4.5cm} >{\centering\arraybackslash}X >{\centering\arraybackslash}X >{\centering\arraybackslash}X >{\centering\arraybackslash}X}
\toprule
Method&732&366&183&92 \\
\midrule
D1:Dual TokenMix&77.07&\textbf{76.22}&\textbf{75.50}&71.48 \\
D2:TokenMix and Dropout Divergent&77.18&75.58&74.51&70.09 \\
D3:Dual TokenMix with Dropout&\textbf{77.35}&76.12&75.40&\textbf{72.90} \\
D4:Dual TokenMix and Single Dropout&77.28&75.77&75.34&70.41 \\
\bottomrule
\label{branch}
\end{tabularx}
\end{table}

\begin{figure}[htbp]
\centering
\setlength{\abovecaptionskip}{-10pt} % 设置图片与标题之间的距离
\centerline{\includegraphics[width=\columnwidth]{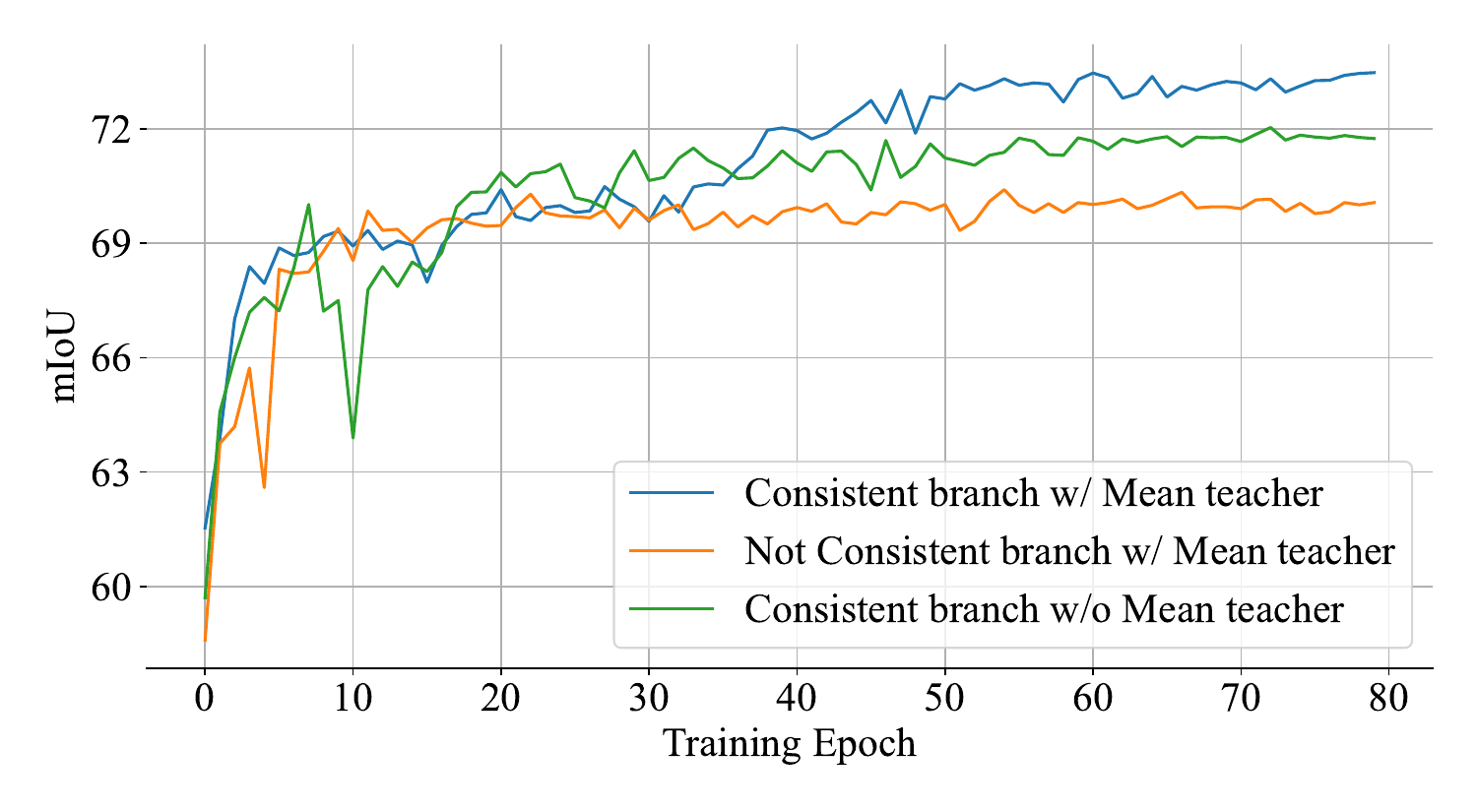}}
\caption{Comparison of consistency regularization for optimization stability on Pascal (92 Labels).}
\label{stable}
\end{figure}

\begin{figure}[htbp]
\centering
% 第一个图片
\begin{minipage}[b]{0.48\linewidth} % 占页面宽度的 48%
\centering
\includegraphics[width=\linewidth]{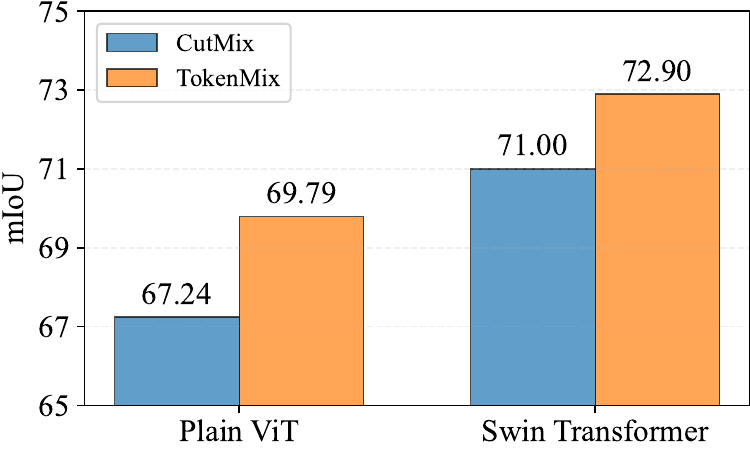}
\caption{Effectiveness of TokenMix in different architectures on Pascal.}
\label{fig:model}
\end{minipage}
\hfill % 在两个 minipage 之间添加水平间距
% 第二个图片
\begin{minipage}[b]{0.48\linewidth} % 占页面宽度的 48%
\centering
\includegraphics[width=\linewidth]{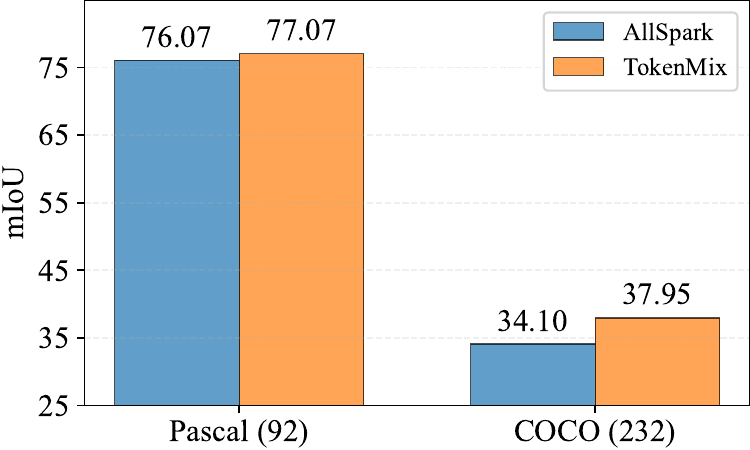}
\caption{Effectiveness of TokenMix in larger models.}
\label{fig:larger model}
\end{minipage}
\end{figure}

% \begin{table}[htbp]
% \caption{Comparison of different methods with different models on Pascal Dataset with 92 labels}
%   \renewcommand{\arraystretch}{0.8}
% \centering
% \begin{tabularx}{1\linewidth}{p{3cm} >{\centering\arraybackslash}X >{\centering\arraybackslash}X}
% \toprule
% Method&CutMix&TokenMix \\
% \midrule
% SETR\cite{31}&69.41&72.52\\
% Segmenter\cite{32}&67.24&71.00 \\
% Swin Transformer\cite{swin}&69.79&72.90\\
% \bottomrule
% \label{table_model}
% \end{tabularx}
% \end{table}

\begin{table}
  \caption{Effect of confidence threshold $\rho$ and momentum decay factor $\theta$ on Pascal (92 labels).}
  \label{parameter}
  \centering
  \renewcommand{\arraystretch}{0.9}
  % 定义新的列类型 C，用于居中对齐
  \newcolumntype{C}{>{\centering\arraybackslash}X}
  \begin{tabularx}{\linewidth}{@{}C|CCCCC@{}}
    \toprule
    $\rho$ & 0.0 & 0.5 & 0.9 & 0.95 & 0.99\\
    mIoU & 65.87 & 70.47 & 71.53 & 72.9 & 69.25\\
    \midrule
    $\theta$ & 0.0 & 0.5 & 0.9 & 0.99 & 0.999\\
    mIoU & 72.04 & 71.80 & 71.56 & 73.48 & 72.90\\
    \bottomrule
  \end{tabularx}
\end{table}

\begin{figure}[htbp]
\centering
\setlength{\abovecaptionskip}{0pt} % 设置图片与标题之间的距离
\centerline{\includegraphics[width=\columnwidth]{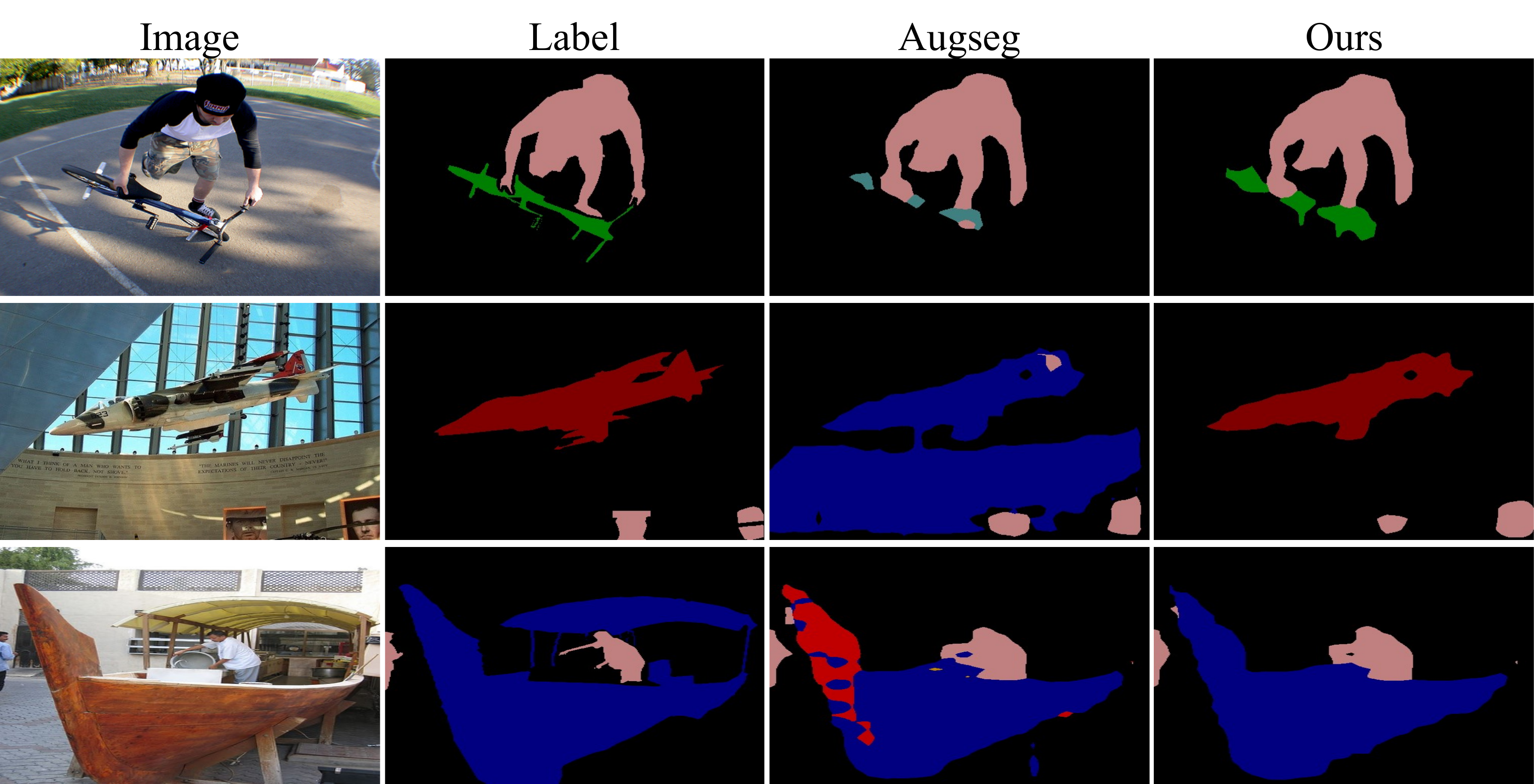}}
\caption{Qualitative comparisons on Pascal (92 Labels).}
\label{visual}
\end{figure}

% \begin{table}[htbp]
% \caption{Comparison of different methods with different models on Pascal Dataset with 366 labels}
% \centering
% \begin{tabularx}{1\linewidth}{p{1cm} >{\centering\arraybackslash}X >{\centering\arraybackslash}X >{\centering\arraybackslash}X >{\centering\arraybackslash}X}
% \toprule
% TokenMix&CorrMatch&RankMatch&PrevMatch&UCCL \\
% \midrule
% \ding{56}&74.60&74.59&75.38&74.59\\
% \ding{52}&75.60&75.24&76.02&75.59 \\
% \bottomrule
% \label{table_model}
% \end{tabularx}
% \end{table}

\vspace{-5pt}

\subsection{Ablation Studies}
\textit{Different Augmentations:} We compare TokenMix with other augmentation methods in Table \ref{augmentation}, where our TokenMix achieves the best results. Combining TokenMix with other augmentations may reduce performance due to excessive transformations disrupting the original semantic structure. Note that \cite{tokenmix} also presents token-level exchange (denoted as TokenMix*). However, it is designed for fully supervised image classification and used for pre-training models that are fine-tuned for fully supervised semantic segmentation, which our TokenMix is specifically designed for S4. Besides, Our TokenMix randomly swaps patches within a single block. In contrast, TokenMix* swaps patches within several small blocks, which introduces much inconsistent semantic information, making it difficult for the model to correctly model the true distribution of unlabeled data in S4. Table \ref{augmentation} shows that our TokenMix significantly outperforms TokenMix* in S4.

\textit{Integration Into Other Methods:} TokenMix can be seamlessly integrated into other SOTA methods to enhance performance further, as shown in Table \ref{integration}.

\textit{Different Branch Designs:} We evaluate four branch designs: D1 (two branches with TokenMix), D2 (one branch TokenMix, one Dropout), D3 (our method: two branches with TokenMix and Dropout), and D4 (two branches TokenMix, one Dropout). Table \ref{branch} shows that D2 and D4, which use different augmentation types per branch, perform worse than D1 and D3, where augmentation types are consistent. This indicates that varying augmentation types produce inconsistent predictions, leading to conflicting gradient signals during parameter updates and increasing the complexity of optimization.

\textit{Optimization Stability:} Figure \ref{stable} shows that consistent branches yield smoother improvements and better final performance. We also compared the mean-teacher method with the dual-branch structure, observing that the mean-teacher method enhances stability and performance. Besides, the dual-branch structure alone outperforms the mean-teacher method, highlighting its significant role in boosting performance.

\textit{Model Compatibility:} TokenMix enhances S4 performance in both standard ViT and hierarchical ViT (Swin Transformer\cite{swin}), as shown in Fig. \ref{fig:model}. Besides, Fig. \ref{fig:larger model} shows our method remains effective with SegFormer-B5, as in AllSpark\cite{14}, proving its applicability to larger models.

\textit{Hyperparameters:} We examine two key hyperparameters in Table \ref{parameter}: momentum decay factor and confidence threshold. Optimal performance is achieved with a confidence threshold of 0.95, balancing noise reduction and information retention. The momentum decay factor should be close to 1 (greater than 0.99) to ensure stable and reliable teacher model outputs.

\textit{Qualitative Comparison:} Qualitative comparisons in Fig. \ref{visual} show that our method correctly identifies the objects (the bicycle in first row and the airplane in second row) and ensures the continuity of the objects (the boat in third row).

\section{Conclusion}
We reconsider data augmentation techniques for ViT-based S4 and propose TokenMix, which introduces new contextual information to help ViT better understand patch relationships. Leveraging ViT's global processing strength, we further introduce a dual-branch framework that combines image and feature augmentation for greater diversity of augmentation. Experiments demonstrate our method's effectiveness, particularly with limited labeled data.

\newpage
\balance

\end{document}